\def\year{2022}\relax
\documentclass[letterpaper]{article}
\usepackage{aaai22}
\usepackage{times}
\usepackage{helvet}
\usepackage{courier}
\usepackage[hyphens]{url}
\usepackage{graphicx}
\usepackage{graphicx}
\usepackage{natbib}
\usepackage{caption}
\DeclareCaptionStyle{ruled}
  {labelfont=normalfont, labelsep=colon, strut=off}
\title{Boosting Anomaly Detection Using Unsupervised Diverse Test-Time Augmentation}
\author{Seffi Cohen, Niv Goldshlager, Lior Rokach, Bracha Shapira}

\date{May 2021}

\usepackage{algorithm}
\usepackage{setspace}
\usepackage[noend]{algpseudocode}
\usepackage{amsmath}
\usepackage{multirow}
\begin{document}

\maketitle

\begin{abstract}
Anomaly detection is a well-known task that involves the identification of abnormal events that occur relatively infrequently. Methods for improving anomaly detection performance have been widely studied. However, no studies utilizing test-time augmentation (TTA) for anomaly detection in tabular data have been performed.
TTA involves aggregating the predictions of several synthetic versions of a given test sample; TTA produces different points of view for a specific test instance and might decrease its prediction bias.
We propose the Test-Time Augmentation for anomaly Detection (TTAD) technique, a TTA-based method aimed at improving anomaly detection performance. TTAD augments a test instance based on its nearest neighbors; various methods, including the k-Means centroid and SMOTE methods, are used to produce the augmentations. Our technique utilizes a Siamese network to learn an advanced distance metric when retrieving a test instance's neighbors. Our experiments show that the anomaly detector that uses our TTA technique achieved significantly higher AUC results on all datasets evaluated.

\end{abstract}

\section{Introduction}
Anomalies are observations that do not meet the expected behavior concerning some context or domain.
Hawkins \cite{hawkins1980identification} defined an anomaly as "an observation which deviates so much from the other observations as to arouse suspicions that it was generated by a different mechanism."
There are three broad aspects that anomaly detection approaches can be characterized by: the nature of the input data, the availability of labels, and the type of anomaly \cite{chandola2009anomaly}. The labels define whether an observation is normal or anomalous. Often, it is challenging to obtain labels for real-world problems, because anomalies are rare \cite{10.1145/1229285.1229292}.

Data augmentation means expanding the data by adding transformed copies of a sample. This technique is used in order to improve a model's performance. By creating transformed copies of a sample, the model can "imagine" alterations for the specific sample to increase the model's generalizability thus performing better on unseen instances \cite{shorten2019survey}. The generalizability of a model refers to its performance when evaluated on new and unknown data; overfitting occurs in models with poor generalizability. Thus, data augmentation aims to close the gap between the known data (seen in the training phase) and any future data, by representing a comprehensive set of possible data points \cite{shorten2019survey}. Most data augmentation methods are applied to image data where the typical transformations are flipping, cropping, rotating, and scaling \cite{mikolajczyk2018data}. Usually, data augmentation is performed when the model is being trained \cite{shorten2019survey}, however it can also be utilized during test time.

Test-time augmentation (TTA) is an application of a data augmentation technique on the test set. TTA techniques generate multiple augmented copies for each test sample, predicting each of them and combining the results with the original sample's prediction.
TTA is more efficient than data augmentation in the training phase, since it does not require retraining the model yet preserves its accuracy and robustness \cite{shanmugam2020and}.

Intuitively, when focusing on image data, TTA could produce different points of view by reducing errors for a specific image. One of the advantages of using augmentation to train convolutional neural networks is that it reduces the error without changing the network’s architecture \cite{perez2017effectiveness}. However, in the case of massive training sets like ImageNet \cite{deng2009imagenet}, it could be expensive to train the network with the augmented data. TTA is much more efficient than data augmentation for training, because no retraining of the model is required. According to the independence principle of ensemble methods \cite{rokach2010ensemble}, the augmented instances' diversity is critical for error reduction \cite{melville2005creating}. Several studies have proposed test-time augmentation techniques to improve robustness and accuracy, or estimate uncertainty \cite{wang2019aleatoric, cohen2019certified, sato2015apac}.

No studies, however, have applied TTA for anomaly detection. In this paper, we propose the Test-Time Augmentation for anomaly Detection (TTAD) technique, which addresses this gap. The use of TTA in anomaly detection could generate an augmented "normal" sample that is out-of-distribution and may reduce the anomaly detector's performance.
There is a trade-off between diversity and generating close enough instances, i.e., in-distribution; the key idea of TTAD is to balance this trade-off by avoiding the generation of out-of-distribution augmentations by learning the test-time augmentations based on similar samples. We use the Nearest Neighbors (NN) algorithm to retrieve the closest neighbors of a given test sample.
Most NN models use a simple $L_1$ distance or $L_2$ distance, or another standard distance metric to measure and retrieve the closest data points which are represented as vectors
\cite{weinberger2006distance}.
However, the above mentioned standard distance metrics are naive and cannot adequately capture the underlying properties of the input data \cite{NGUYEN2016805, weinberger2006distance}.
Ideally, NN models should be adapted to a particular problem in terms of the distance metrics. Domeniconi \cite{domeniconi2005large} showed how NN classification performance significantly improved when a learned distance metric was used.

In this paper we also propose an extension of the NN algorithm for selecting the closest neighbors, utilizing a learned distance metric in the NN model.
Siamese neural networks \cite{koch2015siamese} can learn an adaptive distance between two input samples by calculating the distance of their learned embeddings \cite{lewisaugmenting}. As a result, a trained Siamese neural network's forward propagation can be considered as a distance metric in an NN model. In our case of tabular data, the input of a Siamese network accepts a pair of two tabular records. The network then computes a distance metric between the pair's embedding and outputs a scalar representing this learned distance.

We examined different approaches for creating tabular TTAs based on the output (the subset of data) of the NN model.
Our technique produces test-time augmentations on a given test instance, with trained k-Means centroids as synthetic augmentations, in order to produce a diverse set of in-distribution TTAs. We also examine SMOTE
as an augmentation producer. 
 
Our main contributions can be summarized as follows:
\begin{itemize}
\item We present TTAD, a novel TTA technique for tabular anomaly detection tasks.
\item Our technique produces diverse augmentations in a controlled manner by avoiding out-of-distribution augmentations, using an NN algorithm; the augmentation producers generate the required diversity.
\item Our extensive empirical experiments demonstrate that our method outperforms the baselines.  
\end{itemize}

\section{Background \& Related Work}
\subsection{SMOTE}
In SMOTE \cite{chawla2002smote}, each minority class sample is augmented using each of the $k$ nearest neighbors, with a randomly selected number of neighbors. A linear interpolation of existing minority class samples yields the generated samples. Each minority class sample is oversampled as follows:
\begin{align*}
    x_{synthetic} = x + \lambda * | x - x_k |
\end{align*}
where $\lambda$ is a random value between $\{0, 1\}$, $x_k \in A$, and $A$ represents the $k$ nearest neighbors of the minority class sample $x$.
In our work, we utilize the SMOTE oversampling approach to create test-time augmentations.


\subsection{Isolation Forest}
Liu \cite{10.1109/ICDM.2008.17} proposed a method called Isolation Forest (IF) to explicitly identify anomalies instead of profiling normal data points. An isolation forest is built based on decision trees, where partitions are created by randomly selecting a feature and then arbitrarily splitting the value of the selected feature.
Anomalies are those instances that have short average path lengths on the trees, because they are less frequent than regular observations and fewer splits are necessary. In our work, we used the Isolation Forest method as a label propagation algorithm for Siamese network training.

\subsection{Autoencoder-Based Unsupervised Anomaly Detection} Dau \cite{dau2014anomaly} proposed a method that utilizes a replicator neural network, also called an autoencoder, for anomaly detection, which can operate both in one-class and multi-class settings. When training the network to reconstruct only "normal" observations, the assumption is that the trained network should have low reconstruction error on normal samples, because it learned how to replicate the normal samples. In contrast, when handling anomalous samples, the reconstruction error should be higher, because the network is not trained to replicate these samples. We used this anomaly detection strategy to evaluate our approach.

\subsection{Test-Time Augmentation}
Kim \cite{kim2020learning} proposed a dynamic TTA method for images, in which each transformation is adapted to the current sample by learning the predicted loss of transformed images from the training data. This method selects the TTA transformation with the lowest predicted loss for a given image. Thus, the final classification for a given image is calculated by averaging the target network’s average classification outputs over the transformations that have lower predicted losses. We used TTA for anomaly detection, a novel application of TTA that was not used in prior studies.

\begin{figure*}[tb]
    \centering
    \includegraphics[width=1\textwidth]{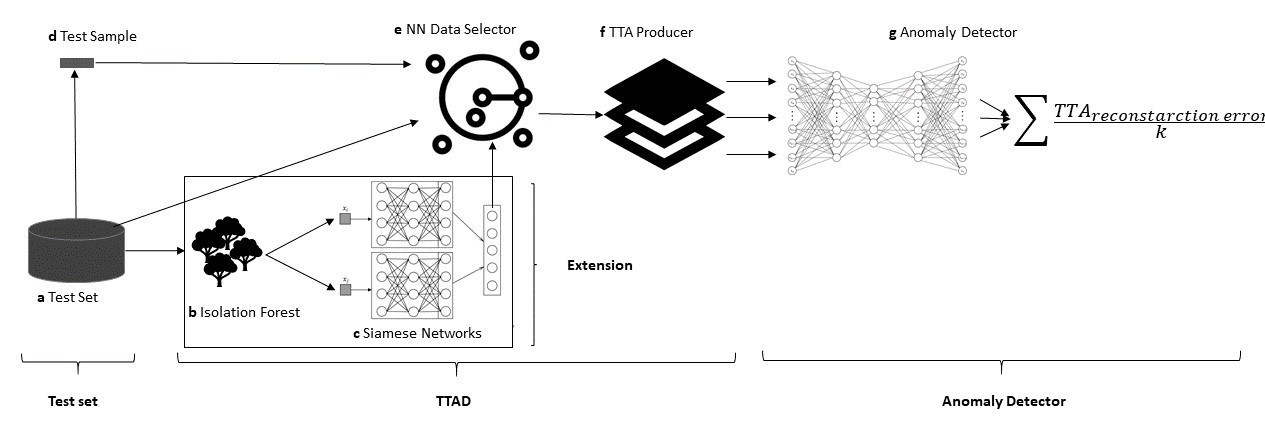}
    \caption{An overview of the TTAD technique. \textbf{a} - The test set. \textbf{b} - An isolation forest is used to pseudo-label the test set. \textbf{c} - A custom distance metric for the nearest neighbor data selector is learned using a Siamese network. \textbf{d} - TTAD is applied for each instance in the test set. \textbf{e} - The data selector component selects a subset of instances for each instance in the test set to serve as a training set for generating augmented instances. \textbf{f} - The augmentation producer component generates diverse augmented instances. \textbf{g} - The instances are scored using an anomaly detector, and the anomaly score is aggregated by the mean.}
    \label{fig:phase1phase2overview}
\end{figure*}

\section{Method}
Figure \ref{fig:phase1phase2overview} shows the proposed TTAD technique, including the entire test phase. TTAD produces diverse augmentations for the test set and combines a balance subset of augmentations for each instance. Then it predicts the augmentations using the anomaly detector.
TTAD is based on two main components. (1) The data selector component, which aims to select a subset of the data to learn the augmentation for each instance in the test set. It enforces similarity between the selected neighbors; and (2) the augmentation producer component, which generates the test-time augmentations and ensures diversity in a controlled manner.

The trade-off between diversity and generating in-distribution instances is balanced by the two components, as shown in Algorithm \ref{alg:our_tta_algorithm}. \\
For each $instance$ in $X_{test}$, the data selector component (lines 1-5) selects a $Subset_{instance}$ of $k$ points in ($X_{train} \cup X_{test}$). The augmentation producer component (line 6) generates $\mathcal{T}$ augmentations using $\mathcal{P}$ as an augmentation method, based on the $Subset_{instance}$. Then the predictions of the $instance$ and the augmentations, which are obtained by the anomaly detector $\mathcal{AD}$, are aggregated (lines 7-8) to produce the final prediction.

The TTAD technique, as presented, is a training-free algorithm, which can easily be extended to a more general setting and achieve computational efficiency. We also propose an extension for the TTAD technique, which utilizes a learned distance metric in the data selector component, as illustrated in Figure \ref{fig:phase1phase2overview} and described in the Component 1 section.




\begin{algorithm}[t]
    \caption{TTAD}
    \label{alg:our_tta_algorithm}
    \textbf{Input:} \\$k$-Number of neighbors; $\mathcal{T}$-Number of augmentations; $\mathcal{P}$-TTA producer; $\mathcal{AD}$-Anomaly detector; $X_{train}$-Training set; $X_{test}$-Test set; $d$-Distance metric
    \begin{algorithmic}[1]\onehalfspacing
        \State $Predictions\gets$ $\emptyset$
        \State $X\gets X_{train} \cup X_{test}$
        \For{$instance$ in $X_{test}$}
            \State Compute distance $d(instance, X)$
            \State \parbox[t]{\dimexpr\linewidth-\algorithmicindent}{$Subset_{instance}\gets$ select set of $k$ points with the smallest distances $d(instance, X)$\strut}
            \State \parbox[t]{\dimexpr\linewidth-\algorithmicindent}{$TTA\gets$ Produce $t$ augmentations based on $Subset_{instance}$ using $\mathcal{P}$\strut}
            \State Obtain $\hat{y}_{instance}$ and $\hat{y}_{TTA}$ using $\mathcal{AD}$
            \State $\hat{y}_{aggregated}\gets$ $\frac{\hat{y}_{instance} \cup \hat{y}_{TTA}}{\mathcal{T} + 1}$
            \State $Predictions\gets$ $Predictions \cup \hat{y}_{aggregated}$
        \EndFor
        \Return $Predictions$
    \end{algorithmic}
\end{algorithm}

A detailed description of TTAD's two components is provided below.

\subsection{Component 1 - Neighbor-Based Data Selector}
This component aims to select a training subset of the data to learn the augmentation for each instance in the test set, based on the $k$ closest instances. We used an NN model to choose the $k$ closest ones.
The data selector component affects the trade-off between diversity and in-distribution by avoiding the generation of out-of-distribution augmentations. The subset of data (the nearest neighbors) selected during this phase is used to generate in-distribution synthetic samples. As a result, the neighbors selected have an impact on the augmentation producer component.

\begin{figure}[tb]
    \centering
    \includegraphics[width=0.5\textwidth]{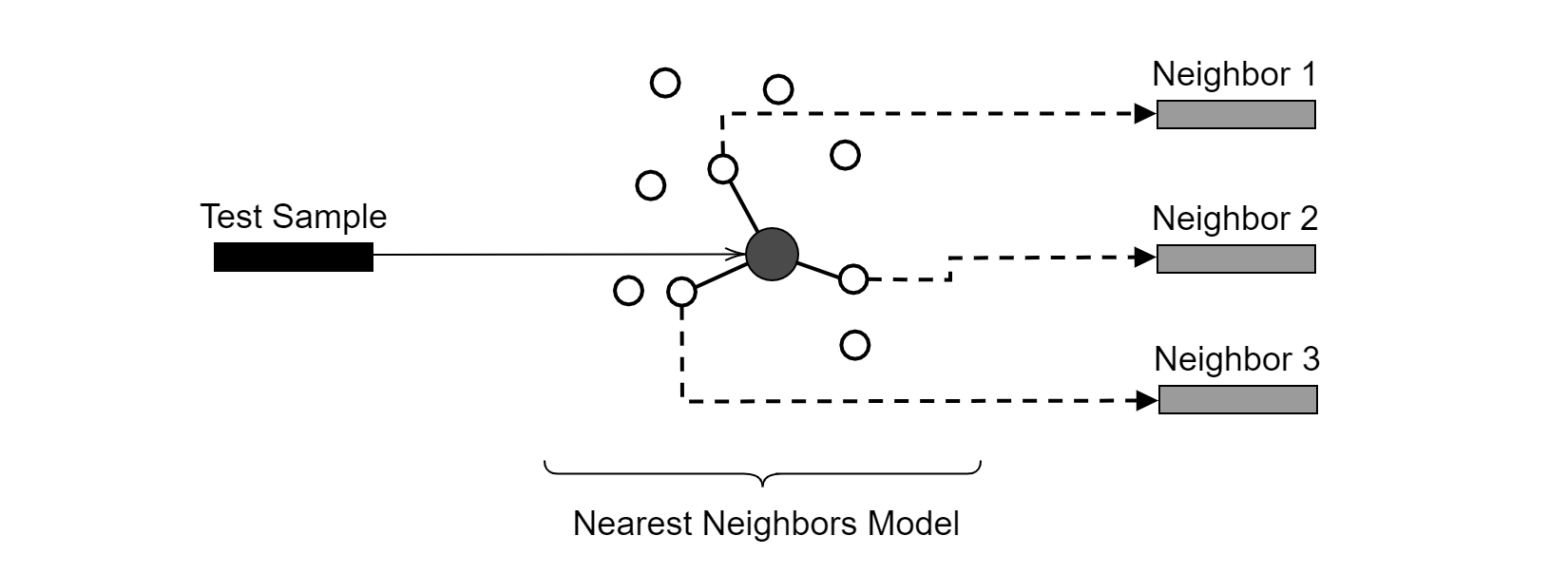}
    \caption{A subset of comparable data is selected for each test instance by the NN model in the data selector component.}
    \label{fig:nearest_neighbors}
\end{figure}

We utilize an NN model that is fitted on both the training and test set data. Given a sample $s$, the NN model retrieves its closest neighbors $\{n_1,n_2,...,n_k\}$ based on a predefined distance metric, where $k$ is the number of nearest neighbors retrieved, as presented in Figure \ref{fig:nearest_neighbors}, where $k=3$.


\subsubsection{Learned Distance Metric}
The basic version of TTAD utilizes the Euclidean distance as the distance metric for identifying the closest neighbors.
However, the Euclidean distance is considered naive, because it does not attempt to learn the data manifold \cite{NGUYEN2016805, weinberger2006distance}.

Therefore, we propose an extension for TTAD's data selector component, utilizing a learned distance metric to overcome the noted limitations of the Euclidean distance.

\begin{figure}[tb!]
    \centering
    \includegraphics[width=0.5\textwidth]{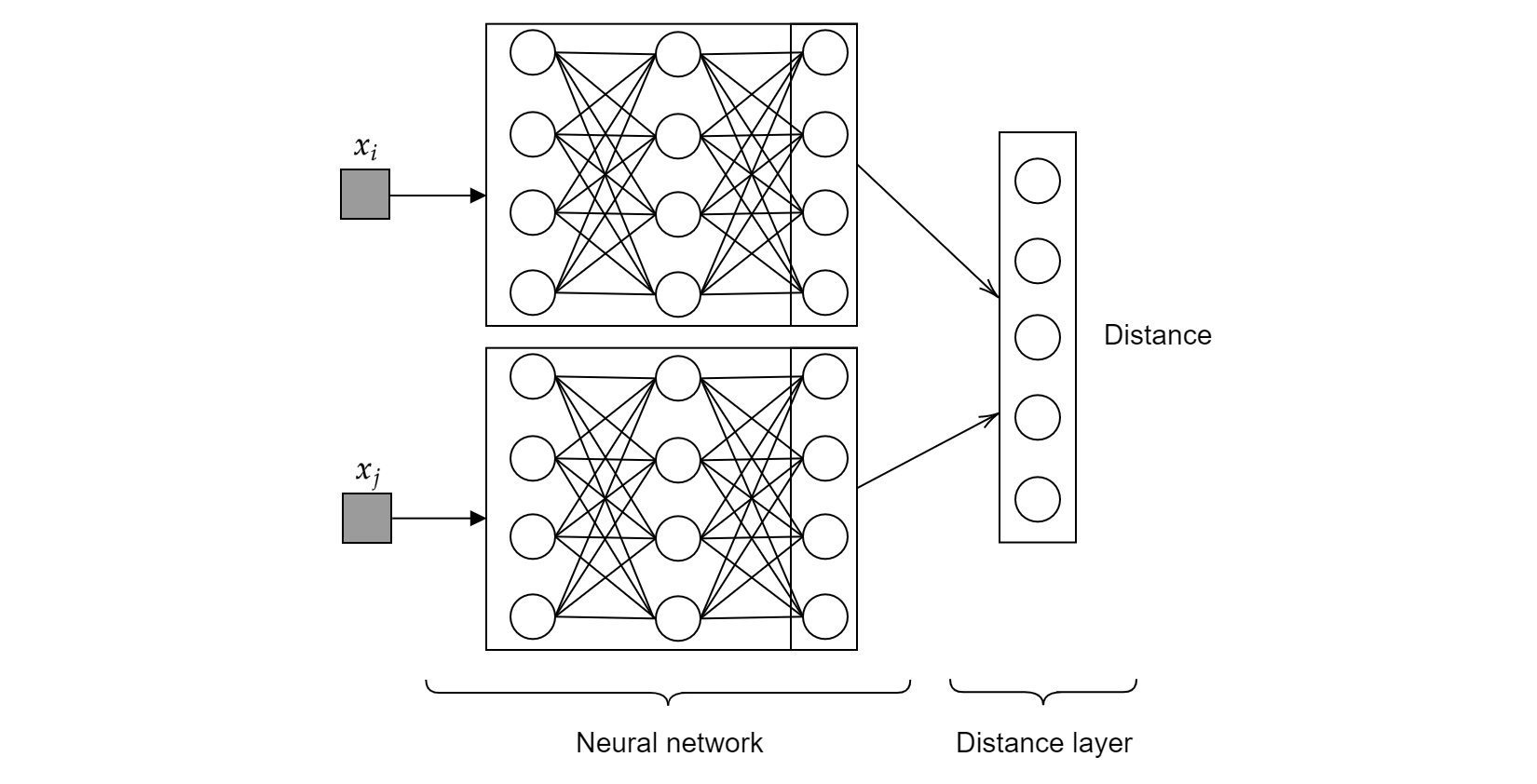}
    \caption{The architecture of the Siamese network: $x_i$ and $x_j$ are the input samples. Their embeddings are obtained by two identical neural networks. The last layer outputs the distance between the input pair embeddings.}
    \label{fig:siamese}
\end{figure}

We use a Siamese network \cite{koch2015siamese} to learn an adapted distance metric between two samples in the context of the given data distribution \cite{lewisaugmenting}. The input of the Siamese network is a pair of samples, and forward propagation on two identical neural networks produces the latent representation of each sample in the pair. The final layer between these two vector representations outputs the distance between the input pair of samples where a distance of zero denotes the closest similarity, and the larger the deviation, the greater the dissimilarity between them.
The following is a more formal definition:
\begin{align*}
    f_{\theta}(x_i, x_j) = d(f_{\theta}(x_i) - f_{\theta}(x_j))
\end{align*}
where $f_{\theta}$ is the network, parameterized by the model weights $\theta$, $d$ is a distance metric (e.g., Euclidean distance), and $x_i$ and $x_j$ are the input samples in a specific pair. Figure \ref{fig:siamese} shows how the final distance between two samples is calculated. First, the two latent representations are obtained by propagating each sample on an identical neural network. Then the distance layer computes the difference between the two representations.

Training a Siamese network requires input of similar and dissimilar pairs, i.e., pairs of inputs from the same class and pairs of inputs from different classes (normal or anomalous in our task). The loss function is then adjusted to a binary classification problem.

However, since TTAD is an unsupervised technique, we cannot use labels to determine if a sample is anomalous or not.
Li \cite{li2019improve} proposed a pseudo-labeling technique for anomaly detection called Outlier Score Propagation (OSP) which uses an IF. We used the same idea to assign a class, i.e., anomalous or normal, for each sample, to create the pairs needed to train the Siamese network.

We assume that using a learned distance metric obtained from the forward propagation of a trained Siamese network will help retrieve more suitable neighbors and thus will produce better test-time augmentations and yield better performance. 

The output of this component is data that will be fed to the augmentation producer. 

\subsection{Component 2 - Augmentation Producer}
This component generates the test-time augmentations based on the selected data from component 1. We present and compare several ways to augment a given test sample using nearest neighbors, including the use of k-Means centroids as synthetic augmentations and SMOTE.


\subsubsection{K-Means TTA Producer}
\begin{figure}[tb!]
    \centering
    \includegraphics[width=0.5\textwidth]{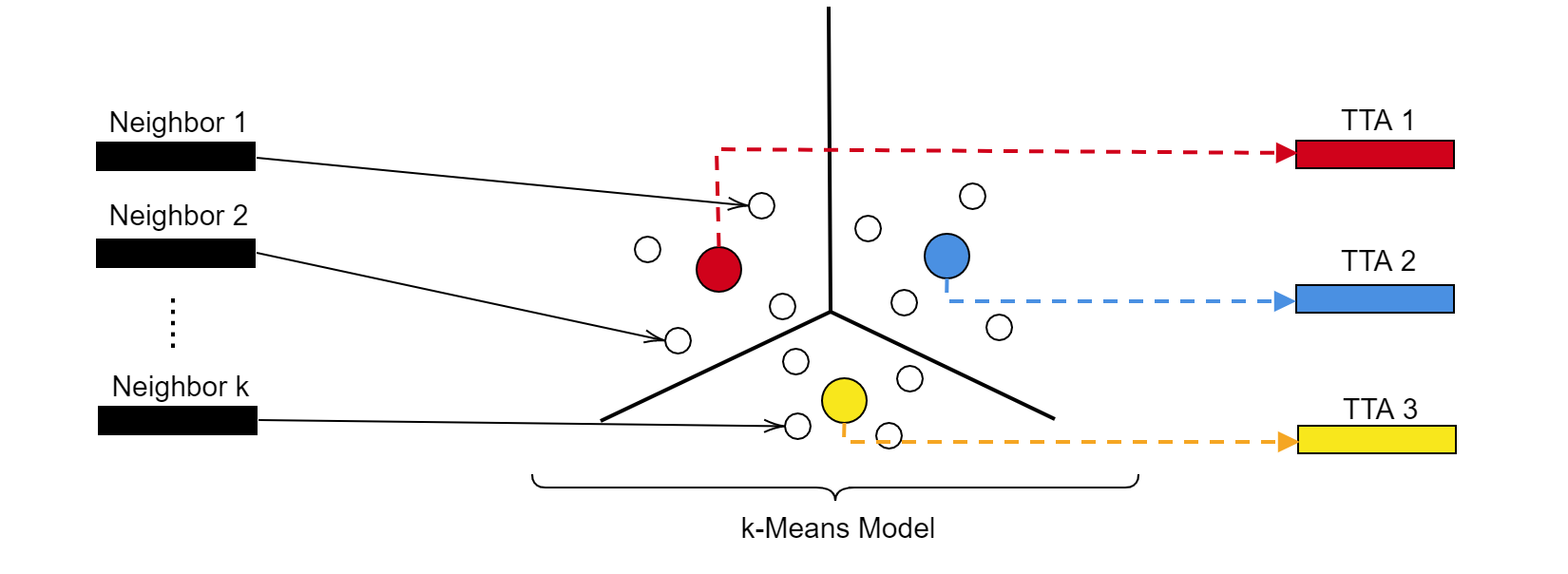}
    \caption{Generation of diverse augmentations using the k-Means model’s centroids on data selected in component 1.}
    \label{fig:kmeans}
\end{figure}

The centroids of a trained k-Means model can be considered as TTAs. During each iteration of a k-Means algorithm, new coordinates are calculated for each cluster's centroid. Finally, after the algorithm’s convergence, the centroids represent the middle positions of the clusters that represent the given training data. Thus, by fitting a k-Means model with a given test sample's nearest neighbors, we take the centroids of the clusters as TTAs. Figure \ref{fig:kmeans} presents the aforementioned procedure.

This approach introduces diversity in the augmentations. The centroids created by the k-Means algorithm enforce $k$ diverse data points, representing the properties of $k$ subsets of the given data to the greatest extent possible. Such diversity enforcement might improve the generalization, which often leads to error reduction \cite{melville2005creating}.

Using trained k-Means centroids as synthetic samples is quite similar to using SMOTE. Both methods use the feature space to produce new samples and consider the $k$ nearest neighbors. Trained k-Means centroids are data points computed using all of the cluster samples, while in SMOTE, the new data points are obtained using a random neighbor, with an arbitrary gap in the distance between the random neighbor and the oversampled point.

\subsubsection{Oversampling TTA Producer}
We also chose the SMOTE oversampling method
as a TTA producer. Usually, oversampling is applied to cope with imbalance between classes in the dataset and reduce overfitting. 
However, we are are not interested in balancing the data but rather in the synthetic data generated for balancing. 
Thus, we only apply the relevant part of the SMOTE algorithm - the part that creates a synthetic sample. SMOTE
creates new instances by randomly interpolating new data points from existing ones. In our case, the existing instances are the given test sample's nearest neighbors. SMOTE
can create new synthetic samples which will be the test-time augmentations.

\begin{figure}[tb!]
    \centering
    \includegraphics[height=5cm]{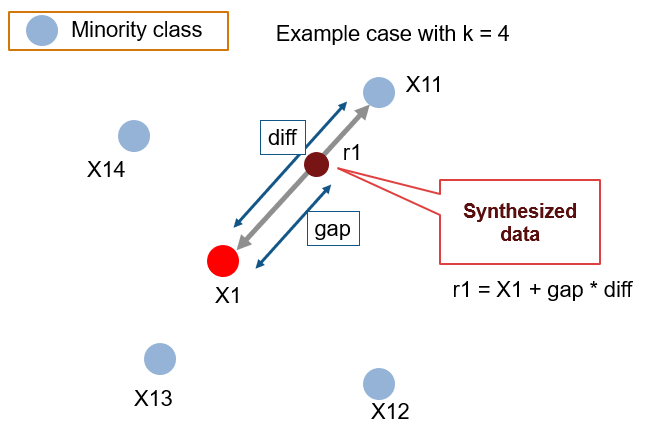}
    \caption[SMOTE FIGURE]{Producing synthetic samples with SMOTE by randomly interpolating new points from existing instances.\footnotemark}
    \label{fig:smote}
\end{figure}

Using k-Means centroids as augmentations produces controlled diversity but may generate augmentations that do not lie in good areas of the data manifold. As a result, we found SMOTE a suitable augmentation producer, because it directly interpolates between two samples to produce augmentations. A visual illustration of SMOTE's procedure is shown in Figure \ref{fig:smote}.

TTAD's output is then fed to an anomaly detector which produces the predictions. Then these predictions are aggregated with the prediction of the input test instance. 

\section{Experiments}

We conduct several experiments to assess TTAD's ability to improve anomaly detection. We examined the basic (Eucledean distance) and advanced (Siamese network) similarity metrics for the data selector component and two augmentation methods for the augmentation producer component: k-Means, SMOTE.

The experiments are performed on eight benchmark tabular anomaly detection datasets, using the proposed TTAD technique on all data selector and data producer combinations, namely:

- Euclidean-NN selector with all producers:
(i)  Euclidean-NN with SMOTE producer (TTAD-ES), 
(ii) Euclidean-NN with k-Means producer (TTAD-EkM)

- Siamese-NN selector with all producers:
(iiii) Siamese-NN with SMOTE producer (TTAD-SS), 
(iv) Siamese-NN with k-Means producer (TTAD-SkM). 
All of the proposed combinations above are compared with two baselines, which are described below.

\footnotetext{\url{https://github.com/minoue-xx/Oversampling-Imbalanced-Data}.}
\subsection{Datasets}
All datasets are available at the Outlier Detection Data Sets repository (ODDS\footnote{http://odds.cs.stonybrook.edu/}) \cite{Rayana:2016}. The ODDS is a public repository of benchmark tabular datasets for anomaly detection. The specific details about each dataset are presented in Table \ref{table:datasets}.
The ODDS collection aims at providing datasets from different domains, and we are interested in evaluating the domain generalizability of TTAD, so it is a suitable choice.

\begin{table}[hbt!]
\renewcommand{\arraystretch}{1.5}
\caption{Description of the ODDS datasets}
\begin{tabular}{c|ccc}
\hline
Dataset          & \#Samples & \#Dim & Outliers(\%) \\ \hline
Yeast            & 1364      & 8     & 4.7          \\
Seismic         & 2584      & 11    & 6.5          \\
Vowels           & 1456      & 12    & 3.4          \\
Annthyroid          & 7200     & 6     & 7.42            \\
Satellite        & 6435      & 36    & 32           \\
Cardiotocography & 1831      & 21    & 9.6          \\
Mammography      & 11183     & 6     & 2.32         \\
Thyroid          & 3772      & 6     & 2.5         
\end{tabular}
\label{table:datasets}
\end{table}

\subsection{Baselines}

We compare the TTAD technique to two baselines - a vanilla test phase, i.e., without any augmentations (w/o TTA), and a TTA method that creates augmentations with random Gaussian noise (GN-TTA) \cite{kim2020learning}.

Since no existing studies applied TTA to anomaly detection, we could not compare our suggested methods to any methods proposed in previous research.  



\begin{table*}[tb!]
\caption{AUC results on the evaluated datasets and methods. All of the TTAD's combinations are set with $k=10$ and $\mathcal{T}=7$. The first two rows are the baselines.}
\renewcommand{\arraystretch}{2.2}
\resizebox{\textwidth}{!}{\begin{tabular}{c|llllllll}
\hline
Method                     & \multicolumn{1}{c}{Cardio} & \multicolumn{1}{c}{Mammography} & \multicolumn{1}{c}{Satellite} & \multicolumn{1}{c}{Seismic} & \multicolumn{1}{c}{Annthyroid} & \multicolumn{1}{c}{Thyroid} & \multicolumn{1}{c}{Vowels} & \multicolumn{1}{c}{Yeast} \\ \hline
w/o TTA                    & 0.956$\pm$0.020             & 0.815$\pm$0.040                  & 0.835$\pm$0.017                & 0.721$\pm$0.041      & 0.762$\pm$0.058     & 0.975$\pm$0.019              & 0.743$\pm$0.139             & 0.735$\pm$0.125            \\
GN-TTA             & 0.618$\pm$0.257             & 0.554$\pm$0.081                  & 0.702$\pm$0.044                & 0.406$\pm$0.138               & 0.564$\pm$0.055              & 0.637$\pm$0.096              & 0.556$\pm$0.305             & 0.522$\pm$0.168            \\ \hline
TTAD-ES            & 0.963$\pm$0.018             & 0.820$\pm$0.044                  & \textbf{0.863$\pm$0.020}       & 0.724$\pm$0.041               & 0.711$\pm$0.060              & 0.960$\pm$0.044              & 0.595$\pm$0.148             & 0.805$\pm$0.110            \\
TTAD-SS              & 0.956$\pm$0.020             & 0.837$\pm$0.042                  & 0.855$\pm$0.022                & 0.725$\pm$0.029               & 0.756$\pm$0.063              & 0.974$\pm$0.026              & 0.692$\pm$0.120             & 0.787$\pm$0.093            \\
TTAD-EkM          & \textbf{0.972$\pm$0.016}    & 0.828$\pm$0.042                  & 0.859$\pm$0.020                & \textbf{0.729$\pm$0.041}               & 0.720$\pm$0.065     & 0.971$\pm$0.030              & 0.693$\pm$0.140             & 0.814$\pm$0.096            \\
TTAD-SkM            & 0.966$\pm$0.018             & \textbf{0.840$\pm$0.043}         & 0.853$\pm$0.022                & 0.726$\pm$0.038               & \textbf{0.763$\pm$0.061}     & \textbf{0.977$\pm$0.025}     & \textbf{0.750$\pm$0.123}    & \textbf{0.823$\pm$0.081}   \\ \hline
\end{tabular}}

\label{tab:AUC_results}
\end{table*}

\subsection{Experimental Setup}

\subsubsection{Estimator}
The base estimator used to detect anomalies is an autoencoder trained only on normal data. At test time, we can easily calculate a given test sample's reconstruction error and predict whether it is anomalous or not.
The reconstruction error output from the autoencoder is the anomaly score of the input test sample.
We utilize an autoencoder with mirrored encoder and decoder architectures. The autoencoder has the same architecture for all experiments and is tuned to have one hidden layer with 64 neurons, where the latent space dimension is 16. The input and output dimensions are dependent on the dataset (number of features). The activation function in all hidden layers is ReLU, and for the output layer, we use sigmoid.

\subsubsection{Siamese Network}
The architecture of the Siamese network used in our experiments is slightly different from the standard Siamese CNN. We deal with tabular data, and therefore we replace the convolutional layers with fully connected ones. We use an architecture of two hidden layers where the first layer has 32 neurons and the second has 64. The activation function used in the hidden layers is ReLU, while for the output layer, we use sigmoid. For training, the final distance layer between the two latent representations is performed using the $L_1$ distance. Eventually, when using the trained Siamese network as a distance metric, we replace the distance metric with the cosine similarity \cite{lee2021local}:
\begin{align*}
    similarity(x, y) = \frac{\sum^{k}_{i=1}x_{i}y_{i}}{\sqrt{\sum^{k}_{i=1}x_i}\sqrt{\sum^{k}_{i=1}y_i}}
\end{align*}
where $x$ and $y$ are vectors with $k$ dimensions.

The isolation forest, used for pseudo-labeling to create the training set for the Siamese network as described in the Method section, is built using 200 trees and without bootstrapping.

\subsubsection{Evaluation Metric} We evaluate the performance of the compared methods using the area under the receiving operator curve (AUC) metric, which is agnostic to the anomaly score threshold and is commonly used in anomaly detection \cite{li2019improve, lim2018doping}.

We perform 10-fold cross-validation, and therefore the mean and standard deviation on all splits are reported for each dataset and method. As a result, the stated results are more robust and reliable.

\subsubsection{Experimental configurations and hyperparameters}
In our experiments, we use the same anomaly detection estimator for all of the datasets and methods evaluated. We train the autoencoder for 300 epochs with a batch size of 32.
We use the Adam \cite{kingma2014adam} optimizer with an initial learning rate of $10^{-3}$, while the other parameters remained unchanged.
The loss function used is the mean squared error (MSE) which was also used in \cite{10.1145/2689746.2689747}. In our case, when using an autoencoder, the MSE is the reconstruction error minimized during training.
\begin{align*}
    \mathcal{L}(x, \hat{x}) = MSE =  \frac{1}{m}\sum^{m}_{i=1}(x^{(i)}-\hat{x}^{(i)})^2
\end{align*}
where $m$ is the number of samples, $x_i$ is sample $i$, and $\hat{x}_i$ is the reconstructed sample $i$ produced by the autoencoder.

The Siamese network is trained with an equal number of pairs from the same and different classes, i.e., 50\% of the training pairs are pairs of samples considered by the isolation forest as being in the same class and 50\% that are pairs from different classes. For each dataset, the Siamese network is trained for 10 epochs with a batch size of 64. For training, the Adam optimizer is used with an initial learning rate of $10^{-3}$, while the other parameters remained unchanged, and binary cross-entropy is used as the loss function:
\begin{align*}
    \mathcal{L}(y, \hat{y}) = -\frac{1}{m} \sum^{m}_{i=1} y^{(i)}log\hat{y}^{(i)} + (1-y^{(1)})log(1-\hat{y}^{(i)})
\end{align*}
where $m$ is the number of pairs; $y_i$ is the label of pair $i$, taking the isolation forest predictions as the ground truth; and $\hat{y}_i$ is the prediction of pair $i$ produced by the Siamese network, i.e., same or different class.

Our proposed technique and experiments are implemented using TensorFlow\footnote{https://www.tensorflow.org/} 2.x and RAPIDS\footnote{https://rapids.ai/} with CUDA 10.1. The experiments are run utilizing Nvidia GeForce RTX 2080 Ti with 11G memory and 32G RAM on a CentOS machine.

The benchmark datasets, the anomaly detector, and TTAD's reproducible source code are available online.

\section{Results}

The results for all of the evaluated datasets and methods are summarized in Table \ref{tab:AUC_results}. The results show that our proposed technique, TTAD, outperforms the two baselines on all datasets. It can also be seen that the GN-TTA baseline achieves the worst results by a large margin. Compared to the SMOTE TTA producer, the k-Means TTA producer achieved the best results. Furthermore, the experiments show that the use of the learned distance metric by the TTAD data selectors often yielded better results, where k-Means with the learned distance metric outperformed the baselines on all datasets.

\subsection{Sensitivity Analysis}
Two important hyperparameters in our method are: (1) in the data selector component, the number of neighbors $k$ as a subset of the data for generating the augmentations in the augmentation producer component, and (2) in the augmentation producer component, the number of augmentations $\mathcal{T}$, i.e., the number of synthetic samples, to generate from a given test sample's nearest neighbors retrieved by the data selector component.
We run multiple experiments with different values for the number of neighbors $k \in \{10, 20, 30, 40, 50\}$ and augmentations $\mathcal{T} \in \{4,5,6,7,8\}$ on the Mammography dataset.

\begin{table}[tb]
\caption{The AUC score of the TTAD combinations on the Mammography dataset with $\mathcal{T}=7$ and different numbers of neighbors - $k$.}
\renewcommand{\arraystretch}{2.2}
\resizebox{\columnwidth}{!}{\begin{tabular}{c|lllll}
\hline
\ $k$                & \multicolumn{1}{c}{10} & \multicolumn{1}{c}{20} & \multicolumn{1}{c}{30} & \multicolumn{1}{c}{40} & \multicolumn{1}{c}{50} \\ \hline
TTAD-ES            & 0.820$\pm$0.044           & 0.818$\pm$0.053           & 0.830$\pm$0.051           & 0.803$\pm$0.066           & \textbf{0.849$\pm$0.040}  \\
TTAD-SS              & 0.837$\pm$0.042           & 0.832$\pm$0.049           & 0.844$\pm$0.056           & 0.820$\pm$0.057           & \textbf{0.863$\pm$0.034}  \\
TTAD-EkM          & 0.828$\pm$0.042           & 0.821$\pm$0.050           & 0.829$\pm$0.051           & 0.799$\pm$0.066           & \textbf{0.842$\pm$0.041}  \\
TTAD-SkM            & 0.840$\pm$0.043           & 0.842$\pm$0.048           & 0.849$\pm$0.052           & 0.821$\pm$0.064           & \textbf{0.866$\pm$0.033}  \\ \hline
\end{tabular}}
\label{tab:neighbors_sensitivity}
\end{table}

\begin{table}[tb]
\caption{The AUC score of the TTAD combinations on the Mammography dataset with $k=20$ and different numbers of augmentations - $\mathcal{T}$.}
\renewcommand{\arraystretch}{2.2}
\resizebox{\columnwidth}{!}{\begin{tabular}{c|lllll}
\hline
\ $\mathcal{T}$            & \multicolumn{1}{c}{4} & \multicolumn{1}{c}{5} & \multicolumn{1}{c}{6} & \multicolumn{1}{c}{7} & \multicolumn{1}{c}{8} \\ \hline
TTAD-ES            & 0.831$\pm$0.050          & \textbf{0.838$\pm$0.061} & 0.816$\pm$0.058          & 0.818$\pm$0.052          & 0.814$\pm$0.053          \\
TTAD-SS              & 0.846$\pm$0.046          & \textbf{0.849$\pm$0.064} & 0.834$\pm$0.060          & 0.831$\pm$0.049          & 0.835$\pm$0.041          \\
TTAD-EkM          & 0.832$\pm$0.048          & \textbf{0.841$\pm$0.062} & 0.822$\pm$0.056          & 0.821$\pm$0.050          & 0.824$\pm$0.049          \\
TTAD-SkM            & 0.847$\pm$0.046          & \textbf{0.851$\pm$0.062} & 0.842$\pm$0.059          & 0.841$\pm$0.047          & 0.844$\pm$0.050          \\ \hline
\end{tabular}}

\label{tab:augmentations_sensitivity}
\end{table}

\subsubsection{Sensitivity to Number of Neighbors} In Table \ref{tab:neighbors_sensitivity}, the AUC results are presented when varying $k$ between 10 and 50 in increments of 10, while $\mathcal{T}$ is set at seven. We can see that using 50 neighbors results in the best performance for all of the TTAD producers we examined. The experiments show that increasing the number of neighbors does not always result in a better AUC, except for Siamese k-Means, whose AUC increases as $k$ increases.

\subsubsection{Sensitivity to Number of Augmentations} In Table \ref{tab:augmentations_sensitivity}, the AUC results are presented when changing $\mathcal{T}$ between four and eight in increments of one, while $k$ is set at 20. We can observe that the best results are obtained when $\mathcal{T}=5$. Note that this only applies when using the Mammography dataset with 20 neighbors.

\section{Discussion}
Our results demonstrate that using tabular TTA can help in the detection of tabular data anomalies. Furthermore, the use of a learned distance metric to retrieve the nearest neighbors and producing TTAs from them often helps boost performance.

We can infer that the GN-TTA baseline achieved such poor results because of the trade-off between diversity and in-distribution augmentations. The GN-TTA method likely generated out-of-distribution samples, which scored normal instances as anomalous, dramatically decreasing the AUC score. 

The k-Means TTA achieved the best results. The k-Means TTA produced more diverse augmentations, because the centroids are taken from different clusters and may lie in better areas of the data manifold.

\section{Conclusion}

In this paper, we introduced TTAD, an unsupervised TTA technique for improving performance on tabular anomaly detection tasks. Our technique manages the trade-off between creating in-distribution test-time augmentations and their diversity.
We also presented an adaptive distance metric to retrieve the nearest neighbors for data selector and several approaches for generating augmentations based on the selected data, including SMOTE 
and k-Means centroids.  
Comparisons on various tabular anomaly detection benchmarks demonstrate that our proposed TTAD technique performs better than an inference without TTA and a Gaussian noise TTA baseline.


\bibliography{references.bib}
\end{document}